\documentclass[runningheads]{llncs}
\usepackage{graphicx}

\usepackage{tikz}
\usepackage{tabu}
\usepackage{comment}
\usepackage{amsmath,amssymb} 
\usepackage{color}

\begin{document}
\pagestyle{headings}
\mainmatter
\def\ECCVSubNumber{1}  

\title{Hardware Architecture of Embedded Inference Accelerator and Analysis of Algorithms for Depthwise and Large-Kernel Convolutions} 


\titlerunning{Hardware Architecture of Embedded Inference Accelerator}
%
\author{Tse-Wei~Chen\inst{1} \and
Wei~Tao\inst{2} \and
Deyu~Wang\inst{2} \and
Dongchao~Wen \inst{2}\orcidID{0000-0001-7311-1842} \and
Kinya~Osa \inst{1} \and
Masami~Kato\inst{1}
}
\authorrunning{T. Chen et al.}
%
\institute{Canon Inc.,\\
30-2, Shimomaruko 3-chome, Ohta-ku, Tokyo 146-8501, Japan\\
\email{twchen@ieee.org} \and
Canon Information Technology (Beijing) Co., Ltd. (CIB),\\
12A Floor, Yingu Building, No.9 Beisihuanxi Road, Haidian, Beijing, China}
\maketitle

\begin{abstract}
In order to handle modern convolutional neural networks (CNNs) efficiently, a hardware architecture of CNN inference accelerator is proposed to handle depthwise convolutions and regular convolutions, which are both essential building blocks for embedded-computer-vision algorithms. Different from related works, the proposed architecture can support filter kernels with different sizes with high flexibility since it does not require extra costs for intra-kernel parallelism, and it can generate convolution results faster than the architecture of the related works. 

The experimental results show the importance of supporting depthwise convolutions and dilated convolutions with the proposed hardware architecture. In addition to depthwise convolutions with large-kernels, a new structure called DDC layer, which includes the combination of depthwise convolutions and dilated convolutions, is also analyzed in this paper. For face detection, the computational costs decrease by 30\%, and the model size decreases by 20\% when the DDC layers are applied to the network. For image classification, the accuracy is increased by 1\% by simply replacing $3 \times 3$ filters with $5 \times 5$ filters in depthwise convolutions.

\end{abstract}

\section{Introduction}
\label{sec:introduction}

Deep learning has been widely applied to image processing and computer vision applications. In the field of embedded vision and robotics, it is important to implement convolutional neural networks (CNNs) with low computational costs~\cite{Chen19}. Many researchers propose efficient algorithms to accelerate the deep-learning-based algorithms while keeping the recognition accuracy~\cite{Chollet16,Han19,Howard17,Qin20,Sandler18,Tan19,Tan19_2}.

Chollet proposes a network architecture called Xception, which includes depthwise convolution layers and point-wise convolution layers, to improve the performance~\cite{Chollet16} of image classification. Howard et al. propose a network architecture called MobileNet, which also includes depthwise convolution layers and point-wise convolution layers~\cite{Howard17}, to reduce the computational costs for embedded computing. Unlike the regular convolution layers in CNNs, the numbers of input feature maps and output feature maps in the depthwise convolution layers are the same, and the computational cost of convolution operations is proportional to the number of input feature maps or output feature maps. Sandler et al. propose MobileNetV2, in which the depthwise convolution layers are still one of the basic structures~\cite{Sandler18}. In addition to these works, there are various kinds of architectures utilizing the concept of depthwise convolutions~\cite{Han19,Tan19,Tan19_2}.

Since depthwise convolutions have become common building blocks for compact networks for embedded vision, the hardware engineers also propose different kinds of accelerators and inference engines to implement them efficiently. Liu et al. propose an FPGA-based CNN accelerator to handle depthwise convolutions and regular convolutions with the same computation cores~\cite{Liu19}. The same architecture is used for the depthwise convolutions and the regular convolutions, but it is difficult to increase the hardware utilization of the depthwise convolutions because the inputs of some processing elements are always set to zeros. Su et al. propose an acceleration scheme for MobileNet, utilizing modules for depthwise convolutions and regular convolutions~\cite{Su18}. Similar to Liu's work~\cite{Liu19}, since two separated modules are used for depthwise convolutions and regular convolutions, the hardware resource cannot be fully utilized. Yu et al. propose a hardware system, Light-OPU, which accelerates regular convolutions, depthwise convolutions, and other lightweight operations with one single uniform computation engine~\cite{Yu20}. It efficiently utilizes the hardware by exploring intra-kernel parallelism. However, since the architecture is optimized for $3 \times 3$ filter kernels, the computational time becomes 4 times when the size of filter kernels is increased from $3 \times 3 $ to $5 \times 5$. The hardware cost increases because extra line buffers are required to store the data. In modern CNN architectures, $3 \times 3$ filter kernels are commonly used, but sometimes it is necessary to increase the receptive field by using large-kernel filters \cite{Peng17} or dilated convolutions \cite{Wei18,Wu19} for certain kinds of applications, such as face detection, image classification, and image segmentation. Therefore, a hardware system that can efficiently handle regular convolutions, depthwise convolutions, and large-kernel filters is desired.

The contribution of this paper is twofold. First, a new hardware architecture for embedded inference accelerators is proposed to handle both depthwise convolutions and regular convolutions. The proposed architecture can efficiently handle convolutions with kernels larger than $3 \times 3$, and it can achieve shorter processing time than the related works. The experimental results show that the size of the proposed hardware is 1.97M gates, while the number of parallel processing units for MAC (Multiply-Accumulate) operations is 512. Second, the features of the supported network architectures are analyzed. By replacing regular convolutions with large-kernel depthwise convolutions, we can reduce the computational costs while keeping the accuracy.

The paper is organized as follows. First, in Sec.~\ref{sec:proposed}, the proposed hardware architecture and supported network architectures are introduced. Then, the experimental results are discussed in Sec.~\ref{sec:results}. Finally, the conclusions are given in Sec.~\ref{sec:conclusion}.

\section{Proposed Hardware Architecture and Supported Network Architectures}
\label{sec:proposed}

The proposed hardware architecture can efficiently handle regular convolutions and depthwise convolutions with large-kernel filters. The hardware architecture and the supported network architecture are introduced in this section. The supported network architecture, the solution based on the proposed hardware, the time chart, and the algorithm are introduced in the following subsections.


\subsection{Network Architecture with Two Types of Convolution Layers}
\label{subsec:network}

\begin{figure}
\centering
\includegraphics[width=12cm]{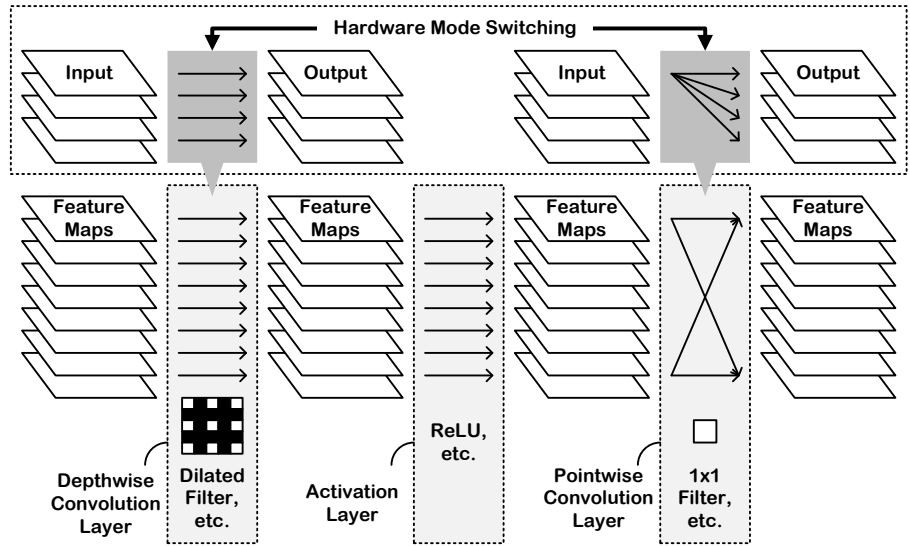}
\caption{Functions of the proposed hardware and an example of the supported network architectures.}
\label{fig:network}
\end{figure}

The upper part of Figure~\ref{fig:network} shows the two modes for regular convolutions and depthwise convolutions supported by the proposed hardware architecture. The main feature of the architectures is the function of hardware mode switching. On the left side, all of the input feature maps for parallel processing are stored in the memory to generate the same numbers of output feature maps. On the right side, only 1 input feature map for parallel processing is stored in the memory to generate multiple output feature maps. When the regular convolution mode is enabled, the intermediate convolution results of multiple input feature maps are accumulated to generate the final convolution results for a single output feature map. When the depthwise convolution mode is enabled, the intermediate convolution results of multiple input feature maps become the final convolution results of multiple output feature maps. In both of the two modes, the hardware architecture can generate the same numbers of convolution results, and the multipliers in the convolution cores are fully utilized. The overhead of the hardware architecture is the buffer to store the data of multiple input feature maps in the depthwise convolution mode. However, large-kernel convolutions can be efficiently handled since the proposed hardware architecture does not require extra costs for intra-kernel parallelism~\cite{Yu20}.

The lower part of Figure~\ref{fig:network} shows an example of the network arhictecture that is focused in this paper. There are 3 layers in this example, the depthwise convolution layer, the activation layer, and the pointwise convolution layer. The depthwise convolution layer may include the filters with kernels larger than or equal to $3 \times 3$. Different from the related works, such as MobileNet~\cite{Howard17}, the depthwise convolution layers also include dilated convolution kernels, which increase the size of the receptive field and increase the accuracy of the inference result for some applications. The combination of dilated convolutions and depthwise convolutions is abbreviated as DDC. The activation layers include functions such as Rectified Linear Unit (ReLU) and quantization functions. The pointwise convolution layer may include the filters with kernel sizes equal to $1 \times 1$ but not limited to $1 \times 1$. When the kernels size is not $1 \times 1$, the pointwise convolutions can also be regarded as regular convolutions.

\subsection{Solution to Depthwise Convolutions and Regular Convolutions}
\label{subsec:solution}

\begin{figure}
\centering
\includegraphics[width=12cm]{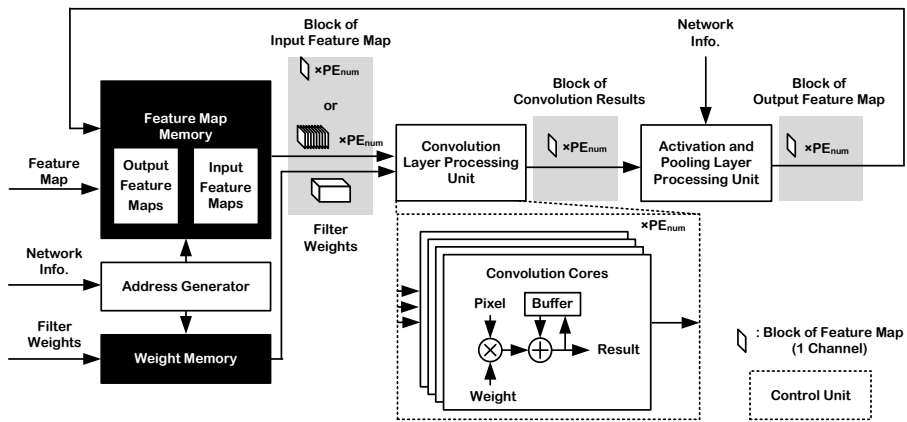}
\caption{Proposed hardware architecture for regular convolutions and depthwise convolutions.}
\label{fig:architecture}
\end{figure}

The proposed hardware architecture is shown in Figure~\ref{fig:architecture}, where the black blocks represent the memory sets, and the white blocks represent the logics. The memory sets include the feature map memory and the filter weight memory. The feature map memory and the weight memory are used to store the input/output feature maps and the filter weights, respectively. The logics include the address generator, the convolution layer processing unit (CLPU), and the activation and pooling layer processing unit (APLPU). The address generator receives the information of network architectures and generates the addresses for the feature map memory and the weight memory. The blocks in the spatial domain of the feature maps are processed sequentially.

The CLPU includes $PE_{num}$ convolution cores to compute the results of convolutions of the input feature maps and the input filter weights in parallel. Each convolution core contains $MAC_{PE}$ sets of multipliers and adders to execute MAC operations.

\subsection{Time Chart}
\label{subsec:timechart}

\begin{figure}
\centering
\includegraphics[width=12cm]{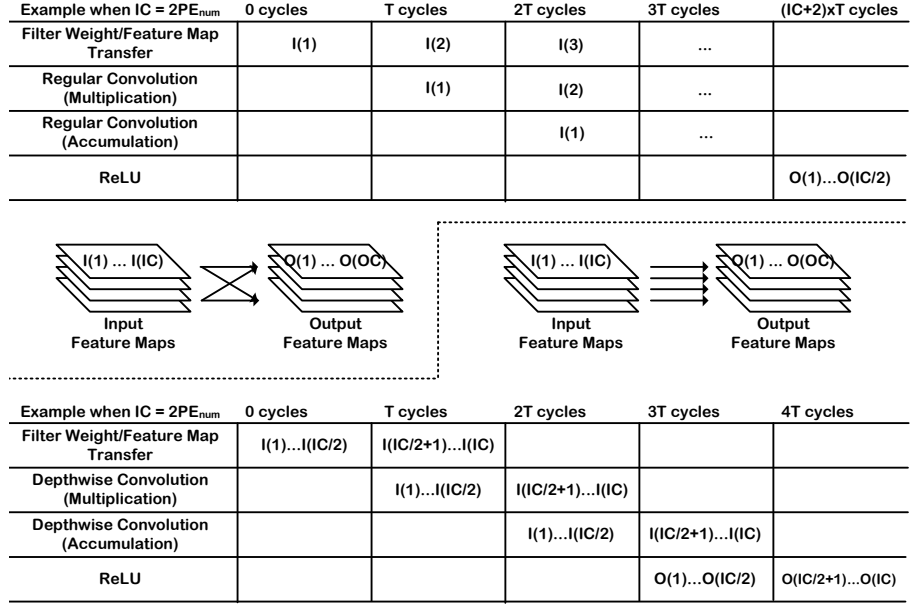}
\caption{Timechart for depthwise convolutions and regular convolutions.}
\label{fig:timechart}
\end{figure}

Figure \ref{fig:timechart} shows a time chart of the hardware architecture, where the operations are executed in pipeline. In this example, we set $IC = 2PE_{num}$ for simplicity. Note that $IC$ represents the number of input feature maps, and $PE_{num}$ denotes the number of convolution cores. The same operations can be applied to regular convolutions and depth-wise convolutions with difference kernel sizes.

The upper part of Figure \ref{fig:timechart} shows an example of regular convolutions. There are $IC$ input feature maps and $OC$ output feature maps in this layer. From cycle 0 to cycle $T$, the 1st input feature map and the corresponding filter weights are transferred. From cycle $T$ to cycle $2T$, the 2nd input feature map and the corresponding filter weights are transferred, and the results of multiplications in the convolution operations based on the 1st input feature maps are computed. From cycle $2T$ to cycle $3T$, the 3rd input feature map and the corresponding filter weights are transferred. The results of multiplications in the convolution operations based on the 2nd input feature maps are computed, and the results of accumulations in the convolution operations based on the 1st input feature maps are computed. From cycle $3T$ to cycle $(IC+2)T$, the results of convolution operations based on the remaining input feature maps are computed. From cycle $(IC+2)T$ to cycle $(IC+3)T$, the results of ReLU based on the convolutions of the 1st to the $IC$-th input feature maps are generated. 

The lower part of Figure \ref{fig:timechart} shows an example of depthwise convolutions. There are $IC$ input feature maps and $OC$ output feature maps in this layer, and the values of $OC$ and $IC$ are the same. From cycle 0 to cycle $T$, the 1st to the $IC/2$-th input feature maps and the corresponding filter weights are transferred. From cycle $T$ to cycle $2T$, the $(IC/2+1)$-th to the $IC$-th input feature maps and the corresponding filter weights are transferred, and the results of multiplications in the convolution operations based on the 1st to the $IC/2$-th input feature maps are computed. From cycle $2T$ to cycle $3T$, the results of multiplications in the convolution operations based on the $(IC/2+1)$-th to the $IC$-th input feature maps are computed, and the results of accumulations in the convolution operations based on the 1st to the $IC/2$-th input feature maps are computed. From cycle $3T$ to cycle $4T$, the results of accumulations in the convolution operations based on the $(IC/2+1)$-th to the $IC$-th input feature maps are computed, and the results of ReLU based on the convolutions of the 1st to the $IC/2$-th input feature maps are generated. From cycle $4T$ to cycle $5T$, the results of ReLU based on the convolutions of the $(IC/2+1)$-th to the $IC$-th input feature maps are generated. 

\subsection{Algorithm}
\label{subsec:algorithm}

\begin{figure}
\centering
\includegraphics[width=12cm]{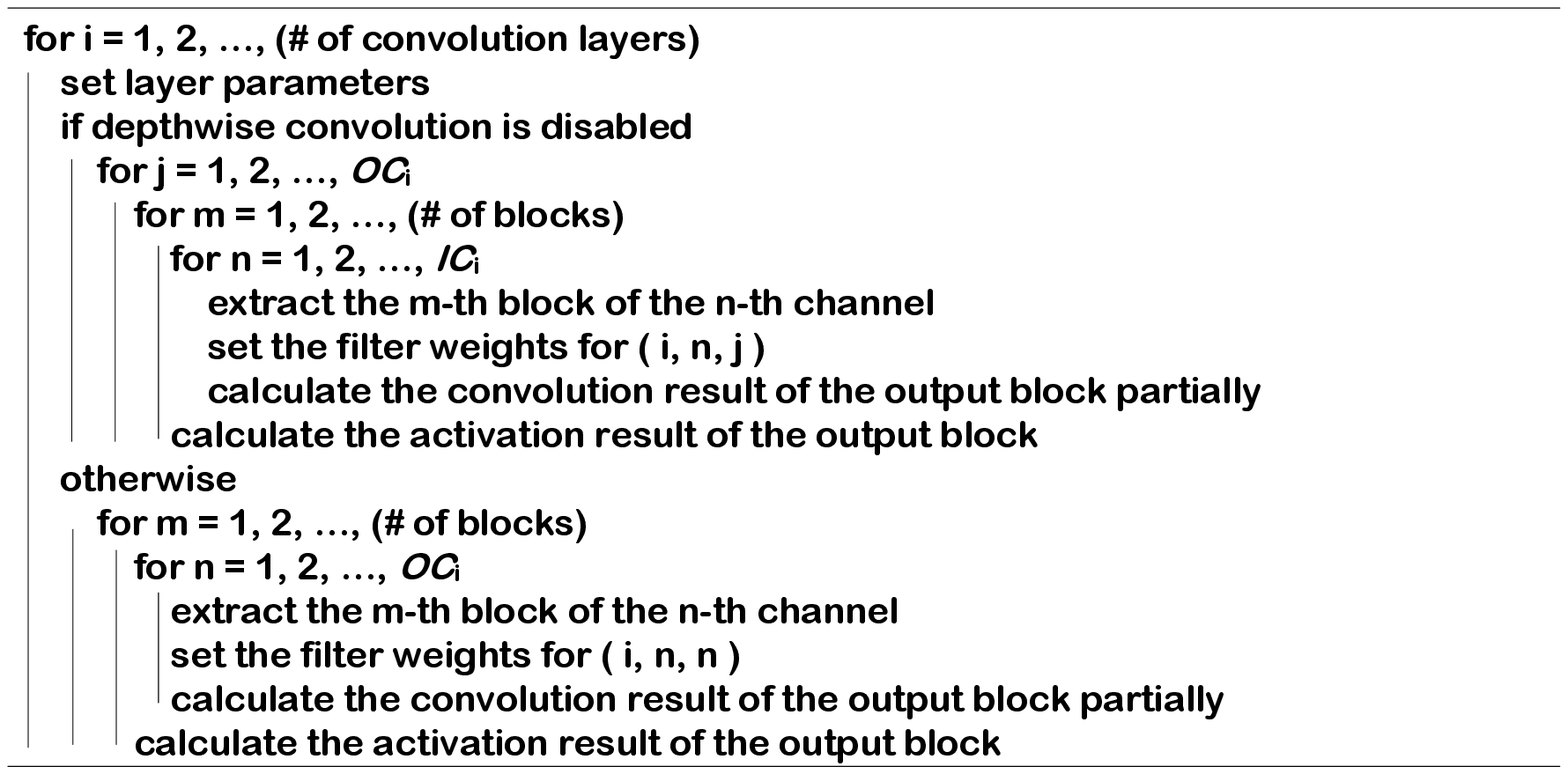}
\caption{Algorithm of the proposed hardware.}
\label{fig:algorithm}
\end{figure}

Figure~\ref{fig:algorithm} shows the proposed algorithms. There are 4 levels in the nested loop structure. In the 1st level of loops, each convolution layer of the networks is processed sequentially, and the parameters of the corresponding layer are set. The mode for depthwise convolutions and regular convolutions is selected in this level.

When the mode for regular convolutions is selected, the operations in the 2nd level to the 4th level of loops are executed. First, in the 2nd level of loops, each output channel of the layer is processed sequentially. Then, in the 3rd level of loops, each block of the output channel is processed sequentially. Finally, in the 4th level of loops, each input channel is processed sequentially. The filter weights of the $i$-th convolution layer, the $j$-th output channel, the $n$-th input channel are set, and the convolution results of $PE_{num}$ output blocks are computed in parallel. 

When the mode for depthwise convolutions is selected, the operations in the 2nd level to the 3rd level of loops are executed. The number of input channels is equal to the number of output channels, and each channel is processed sequentially. The filter weights of the $i$-th convolution layer, the $j$-th output channel, the $j$-th input channel are set, and the convolution results of $PE_{num}$ output blocks are computed in parallel. 

Since the number of loops is reduced when the mode for depthwise convolutions is selected, redundant operations, in which some of filter weights are set to zeros, are not required. The common architecture for the two modes makes the hardware achieve higher utilization than the related works~\cite{Liu19,Su18}.

\section{Experimental Results and Analysis}
\label{sec:results}

The section contains 3 parts. The first part is the comparison of accuracy of face detection and image classification. The second part is the analysis of the computational cost of the proposed hardware architecture. The third part is the comparison of specifications.


\subsection{Comparison of Accuracy}
\label{subsec:comparison}

One of the strengths of the proposed hardware architecture is the function to handle the combination of large-kernel convolutions and depthwise convolutions. To analyze the effectiveness of large-kernel convolutions and depthwise convolutions, the experiments for two kinds of applications are performed. The first application is face detection, and the second application is image classification. Two experiments are performed for both of the applications.

In the first experiment, the accuracy of face detection on the WIDER FACE dataset~\cite{Yang16} is analyzed. The network of RetinaFace~\cite{Deng19} is tested with depthwise convolutions and large-kernel convolutions, and the backbone network is MobileNetV1-0.25~\cite{Howard17}. 

The WIDER FACE dataset~\cite{Yang16} consists of 32,203 images and 393,703 face bounding boxes with a high degree of variability in scale, pose, expression, occlusion and illumination. In order to compare the proposed network architecture with RetinaFace, the WIDER FACE dataset is split into training (40\%), validation (10\%) and testing (50\%) subsets, and three levels of difficulty (i.e. Easy, Medium, and Hard) are defined by incrementally incorporating hard samples. We use the WIDER FACE training subset as the training dataset, and the WIDER FACE validation subset as the testing dataset. The operations in the context module, which is an important structure used to increase the receptive field, include a series of $3 \times 3$ convolutions. Since the feature maps are processed with different numbers of $3 \times 3$ filters consecutively, the effects are similar to applying large-kernel convolutions (e.g. $5 \times 5$, $7 \times 7$) to the feature maps. To show the advantage of the functions of the proposed hardware architecture, some of the operations are replaced with depthwise convolutions and dilated filters. To be specific, two cascaded $3 \times 3$ convolutions are replaced with a single $3 \times 3$ convolution where the dilation rate is 1, and three cascaded $3 \times 3$ convolutions are replaced with a single $3 \times 3$ convolution where the dilation rate is 2.

\begin{table}
\begin{center}
\begin{tabu}{l||ccc}
\tabucline[1pt]{-}
Network & Easy & Medium & Hard\\
\hline
\hline
RetinaFace$^{A}$~\cite{Deng19} (with CM$^{1}$)           & 90.70\% & 87.88\% & 73.50\% \\
RetinaFace$^{A}$ w/o CM$^{1}$                                  & 89.55\% & 86.21\% & 68.83\% \\
RetinaFace$^{A}$ with CM$^{1}$ and DDC$^{2}$                   & 90.28\% & 87.13\% & 73.24\% \\ 
\hline
Quantized RetinaFace$^{A}$~\cite{Deng19} (with CM$^{1}$) & 90.72\% & 87.45\% & 73.56\% \\
Quantized RetinaFace$^{A}$ with CM$^{1}$ and DDC$^{2}$         & 90.32\% & 87.68\% & 73.53\% \\ 
\hline
RetinaFace$^{B}$~\cite{Deng19} (with CM$^{1}$)           & 89.98\% & 87.11\% & 72.01\% \\
RetinaFace$^{B}$ w/o CM$^{1}$                                  & 88.68\% & 84.98\% & 68.56\% \\
RetinaFace$^{B}$ with CM$^{1}$ and DDC$^{2}$                   & 89.60\% & 86.13\% & 71.93\% \\
\hline
Quantized RetinaFace$^{B}$~\cite{Deng19} (with CM$^{1}$) & 89.70\% & 86.91\% & 71.89\% \\
Quantized RetinaFace$^{B}$ with CM$^{1}$ and DDC$^{2}$         & 89.56\% & 86.02\% & 71.75\% \\

\tabucline[1pt]{-}
\end{tabu}
\end{center}
{\small
$^{1}$ CM stands for ``context module."\\
$^{2}$ DDC stands for ``depthwise and dilated convolutions."\\
$^{A}$ The networks are trained from a pre-trained model.\\
$^{B}$ The networks are trained from scratch.\\
}
\caption{Accuracy of RetinaFace~\cite{Deng19} and the proposed network on the
WIDER FACE~\cite{Yang16} validation subset.}
\label{tab:retinaface_accuracy}
\end{table}

The accuracy of RetinaFace~\cite{Deng19} and its variations are shown in Table \ref{tab:retinaface_accuracy}, which includes the proposed network architecture, the depthwise and dilated convolution (DDC) layers. Since the accuracy of the network without the context module is not available in the original paper~\cite{Deng19}, we add an ablation study to verify the effectiveness of the context module. The results show that the context module can increase the accuracy of face detection by about 1\% for the Easy category and the Medium category, and by 4\% for the Hard category. The proposed hardware can support the quantized RetinaFace and its variations. In the quantized networks, the feature maps and the filter weights are both quantized into 8-bit data. It is shown that the quantized versions have similar accuracy with the floating-point versions.

When the filters in the context module are replaced with the DDC layers and pointwise convolutions, the accuracy decreases by less than 1\% for all the categories. The parameter settings are shown in Table~\ref{tab:retinaface_param}. We trained the RetinaFace using the SGD optimizer (momentum at 0.9, weight decay at 0.0005, and batch size of 32) on the NVIDIA Titan Xp GPUs with 12GB memory. The learning rate starts from $10^{-3}$, and is divided by 10 at the 190th and at the 220th epoch. The training process terminates at the 250th epoch.

\begin{table}
\begin{center}
\begin{tabu}{l||c}
\tabucline[1pt]{-}
Backbone & MobileNetV1-0.25\\
\hline
Prediction Levels & $8 \times$, $16 \times$, $32 \times$ down-sampling\\
\hline
Anchor Settings & $16 \times 16$, $32 \times 32$, $8 \times$ prediction layers\\
                & $64 \times 64$, $128 \times 128$, $16 \times$ prediction layers\\
                & $256 \times 256$, $512 \times 512$, $32 \times$ prediction layers\\
\hline
Batch Size & 32\\
\hline
No. of Epochs & 250\\
\tabucline[1pt]{-}
\end{tabu}
\end{center}
\caption{Parameter settings of the experiments for RetinaFace~\cite{Deng19}.}
\label{tab:retinaface_param}
\end{table}

The computational costs and the model sizes of RetinaFace~\cite{Deng19} and the proposed work are shown in Table \ref{tab:retinaface_cost}. When the operations in the context module are replaced with depthwise convolutions and dilated filters, the model size of the context module decreases from 138 KB to 23 KB, and the computational cost of the context module decreases from 708 MACs per input pixel to 119 MACs per input pixel. By applying depthwise convolutions and dilated filters to the network, the total computational costs decrease by about 30\%, and the total model size decreases by about 20\%.

\begin{table}
\begin{center}
\begin{tabu}{lcccc}
\tabucline[1pt]{-}
Network & \multicolumn{2}{c}{Model Size} & \multicolumn{2}{c}{Computational Cost}\\
        & \multicolumn{2}{c}{(Bytes)}    & \multicolumn{2}{c}{(MACs/Input pixel)}\\
\cline{2-5}
        & CM$^{1}$ & Total & CM$^{1}$ & Total \\
\hline
\hline
RetinaFace$^{1}$~\cite{Deng19}    & 138K & 1.12M & 708 & 1,888\\
RetinaFace with CM and DDC$^{2}$ & 23K  & 0.90M  & 119 & 1,298\\ 
\tabucline[1pt]{-}
\end{tabu}
\end{center}
{\small
$^{1}$ CM stands for ``context module."\\
$^{2}$ DDC stands for ``depthwise and dilated convolutions."\\
}
\caption{Computational costs and the model sizes of RetinaFace~\cite{Deng19} and the proposed network.}
\label{tab:retinaface_cost}
\end{table}

In the second experiment, to show the relation between the size of filter kernels and the inference result, the accuracy of image classification on ImageNet~\cite{ILSVRC15} is analyzed. Similar to the first experiment, MobileNetV1-0.25 is used for testing. Since the architecture of MobileNetV1~\cite{Howard17} is composed of depthwise convolution layers, it can be used to evaluate the effectiveness of large-kernel convolutions by simply increasing the kernel sizes of the filters. The accuracy of MobileNetV1 and its variations is shown in Table~\ref{tab:imagenet_accuracy}. There are two networks with dilated convolutions, where the concept of hybrid dilated convolutions is adopted~\cite{Wang17}. When some $3 \times 3$ filters in MobileNetV1 are replaced by $5 \times 5$ filters, the accuracy is increased by more than 1\%. When some of $3 \times 3$ filters are replaced by dilated $3 \times 3$ filters, the network can also achieve higher accuracy than the original architecture.

\begin{table}
\begin{center}
\begin{tabu}{lcc}
\tabucline[1pt]{-}
Network & Top-1 Accuracy & Top-5 Accuracy \\
\hline
MobileNetV1-0.25~\cite{Howard17} & 68.39\% & 88.35\%  \\
MobileNetV1-0.25 with $5 \times 5$ filters & 69.44\% & 89.90\%  \\
MobileNetV1-0.25 with dilated filters (A)$^{1}$ & 68.52\% & 88.61\%  \\
MobileNetV1-0.25 with dilated filters (B)$^{2}$ & 68.98\% & 88.85\%  \\
\tabucline[1pt]{-}
\end{tabu}
\end{center}
{\small
$^{1}$ Dilated convolutions are applied to Layer 14 to Layer 18. The kernel size of filters is $3 \times 3$, and the dilation rates are set to $2,2,2,2,2$ for the 5 layers.\\
$^{2}$ Dilated convolutions are applied to Layer 14 to Layer 18. The kernel size of filters is $3 \times 3$, and the dilation rate are set to $2,3,2,3,2$ for the 5 layers.\\
}
\caption{Accuracy of MobileNetV1-0.25 and the proposed network on ImageNet~\cite{ILSVRC15}.}
\label{tab:imagenet_accuracy}
\end{table}

The first experiment shows that the combination of depthwise convolutions and dilated convolutions keeps the accuracy while reducing the computational costs, and the second experiment shows that the combination of depthwise convolutions and large-kernel convolutions can increase the accuracy.

In brief, these results show the importance of supporting depthwise convolutions and large-kernel convolutions, including dilated convolutions, with the proposed hardware architecture.

\subsection{Computational Time}
\label{subsec:time}

To compare the proposed hardware architecture with the related work, the processing time for regular convolutions and depthwise convolutions is expressed as equations. The processing time (number of cycles) for the regular convolutions is shown as follows.

\begin{align}
  T_{M} = \max \left ( 
  IC \times \left \lceil \frac{IN \times IM }{BW_{FM}} \right \rceil \times
  \left \lceil \frac{OC }{PE_{num}} \right \rceil, 
  IC \times OC \times \left \lceil \frac{X \times Y}{BW_{W}} \right \rceil 
  \right ), 
\end{align}

\begin{align}
  T_{C} = 
  X \times Y \times IC \times \left \lceil \frac{ON \times OM}{MAC_{PE}}  \right \rceil \times 
  \left \lceil \frac{OC}{PE_{num}} \right \rceil ,
\end{align}

where $T_{M}$ and $T_{C}$ are the memory access time and the computational time, respectively. The processing time can be expressed as $\max(T_{M}, T_{C})$, and the definition of parameters in the equations are shown as follows.\\
\\
$IC$: the number of input channels.\\
$OC$: the number of output channels.\\
$IN \times IM$: the size of input blocks.\\
$ON \times OM$: the size of output blocks.\\
$X \times Y$: the size of filters.\\
$BW_{FM}$: the bandwidth to transfer feature maps.\\
$BW_{W}$: the bandwidth to transfer filter weights.\\
$PE_{num} = 8$: the number of processing elements.\\
$MAC_{PE} = 64$: the number of parallel MAC operations implemented in 1 processing element.\\
\\
The memory access time and the computational time for the depthwise convolutions are shown as follows. Similarly, the processing time can be expressed as $\max(T_{M}, T_{C})$. 

\begin{align}
  T_{M} = \max \left ( 
  OC \times \left \lceil \frac{IN \times IM }{BW_{FM}} \right \rceil, 
  OC \times \left \lceil \frac{X \times Y}{BW_{W}} \right \rceil 
  \right ),  
\end{align}

\begin{align}
  T_{C} = 
  X \times Y \times \left \lceil \frac{ON \times OM}{MAC_{PE}}  \right \rceil \times 
  \left \lceil \frac{OC}{PE_{num}} \right \rceil 
\end{align}

All of the processing elements are used for both regular convolutions and depthwise convolutions.

\begin{figure}
\begin{tabular}{cc}
\centering
\includegraphics[width=6cm]{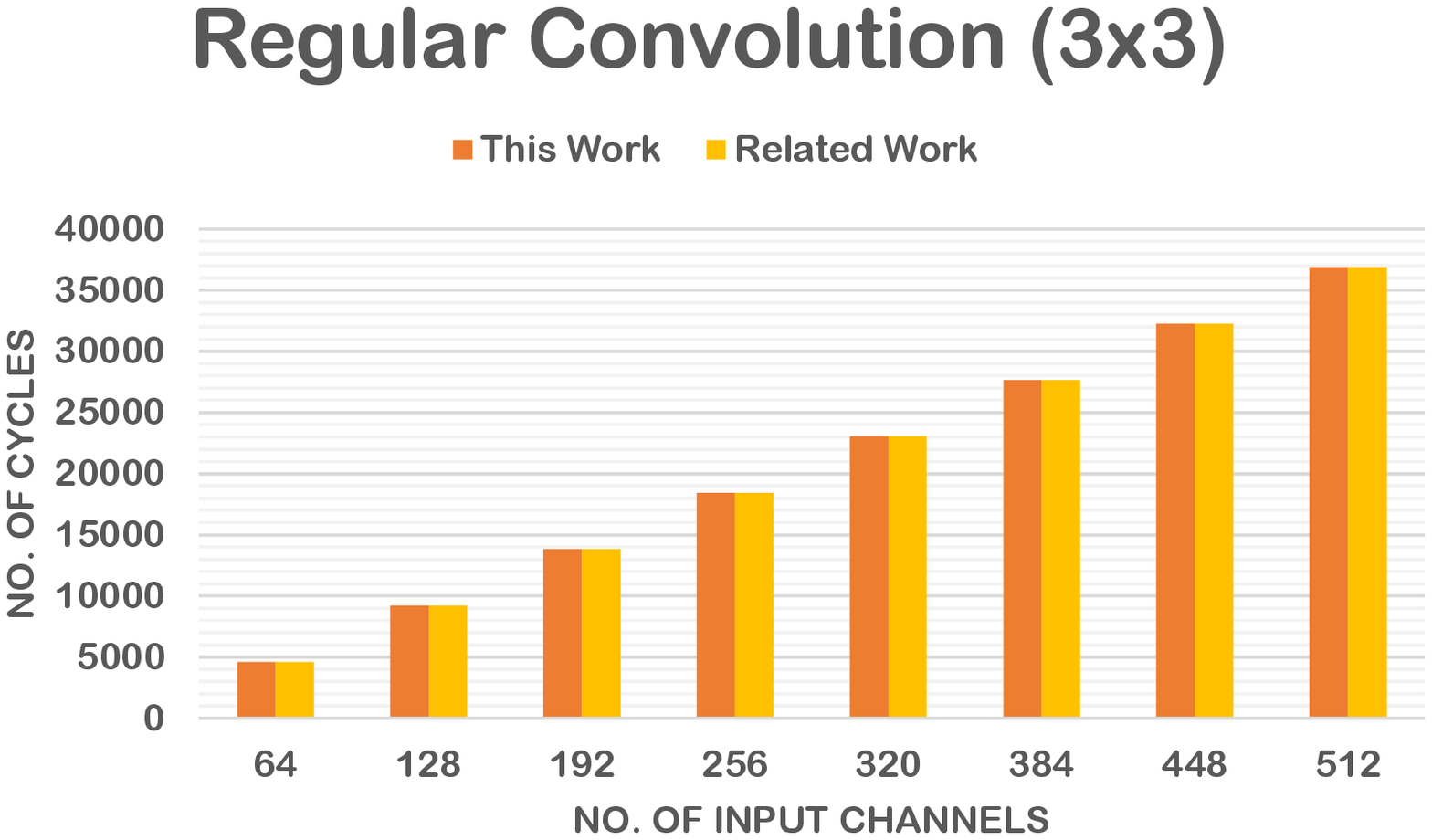} & \includegraphics[width=6cm]{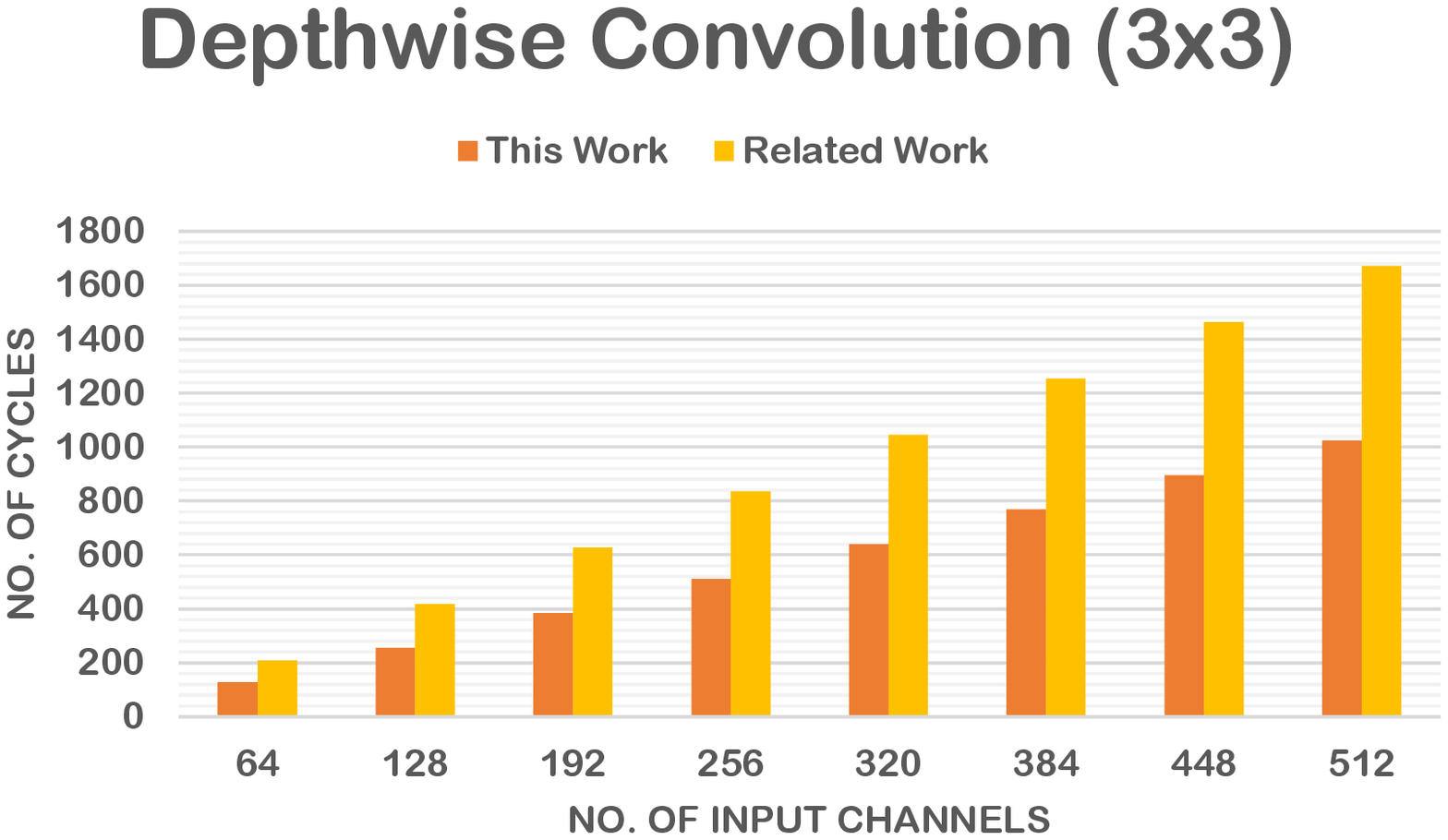}\\
(a) & (b)\\
\includegraphics[width=6cm]{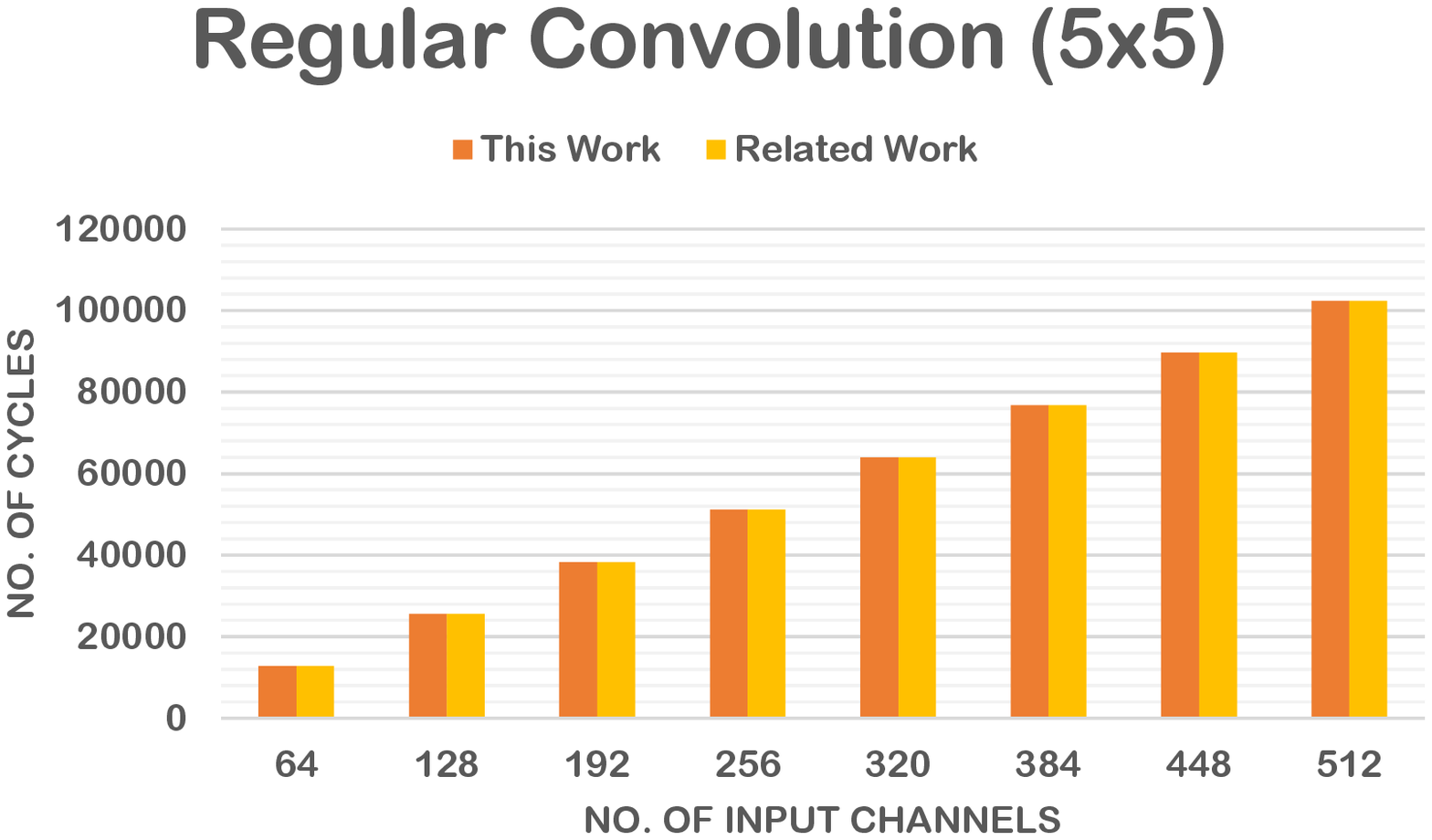} & \includegraphics[width=6cm]{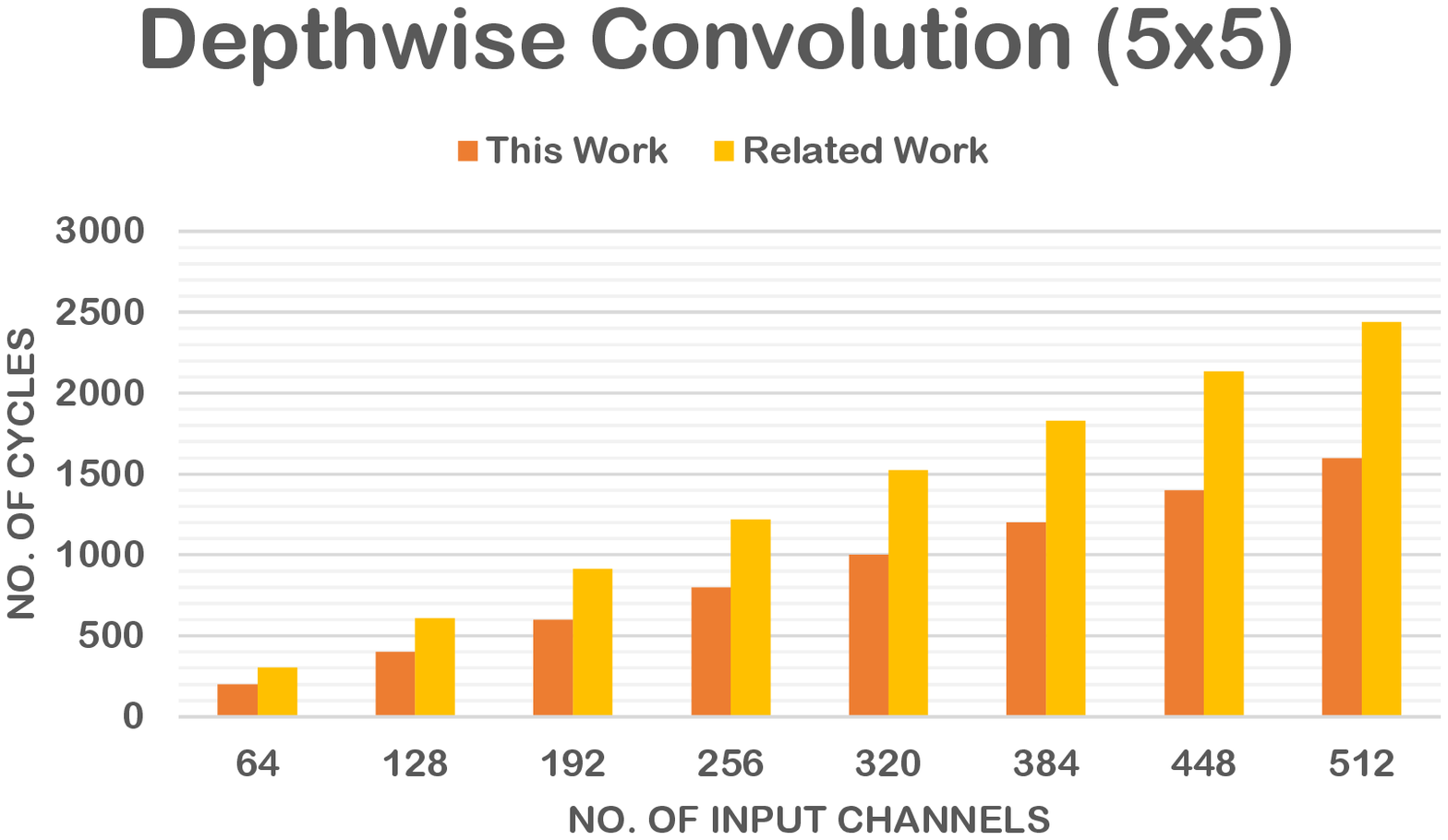}\\
(c) & (d)\\
\end{tabular}
\caption{Comparison of the computational time and the number of input channels with the related work~\cite{Yu20} when the size of filter kernels is (a)(b) $3 \times 3$ and (c)(d) $5 \times 5$.}
\label{fig:comparison}
\end{figure}

In the related work~\cite{Yu20}, the computational costs (the number of cycles) for the convolution operations are also expressed as equations. To compare the efficiency, we set $PE_{num} = 64$, and $MAC_{PE} = 8$ for the related work. For simplicity, $IC$ and $OC$ are set to the same value for both regular convolutions and depthwise convolutions. The results are shown in Fig. \ref{fig:comparison}, where (a) and (b) refer to the cases when the size of filter kernels is $3 \times 3$, and (c) and (d) refer to the cases when the size of filter kernels is $5 \times 5$. For regular convolutions, the computational time of the proposed work and the related work is almost the same when the size of filter kernels is $3 \times 3$ or $5 \times 5$. For depthwise convolutions, the proposed work has shorter computational time than the related work when the size of filter kernels is $3 \times 3$ or $5 \times 5$. In the proposed architecture, since the filter weights and the feature maps can be transferred in parallel, the memory access time can be shortened so that the operations of dilated convolutions achieve faster speed than the related work. Also, since the proposed architecture can process the filter weights sequentially, the computational time is proportional to the size of filter kernels, and the computational efficiency does not decrease when the size of filter kernels becomes large.

\subsection{Specifications}
\label{subsec:specs}

The specifications of the proposed embedded inference accelerator, which is designed to process modern CNNs, are shown in Table~\ref{tab:spec}. The main feature of the proposed architecture is that it can compute depthwise convolutions with large kernels, including the DDC layers mentioned in the previous section.

We use the 28-nm CMOS technology library for the experiments. The resolution of input images is VGA, and the performance is 512 MAC operations per clock cycle for 8-bit feature maps and 8-bit filter weights. As shown in Figure~\ref{fig:architecture}, since regular convolutions and depthwise convolutions can be selected according to the type of layers, the hardware architecture can also process RetinaFace (with MobileNetV1-0.25) efficiently. The gate counts of the CLPU and the APLPU are 1.50M and 0.47M, respectively. The processing speed for RetinaFace, which includes the post-processing algorithm, is 150fps. Compared with RetinaFace, it takes only 83\% of processing time to compute the inference result of the proposed network architecture, DDC layers. Since the proposed hardware can support depthwise convolutions with large filter kernels, the target networks can be switched according to the requirement of processing time.

\begin{table}
\begin{center}
\begin{tabu}{lc}
\tabucline[1pt]{-}
  Gate Count & CLPU$^{1}$: 1.50M \\
  (NAND-Gates) & APLPU$^{2}$: 0.47M\\  
\hline
  Process &  28-nm CMOS Technology\\
\hline
  Clock Frequency &  400 MHz\\  
\hline
  Memory Size & Feature Map Memory: 4,096 KB   \\
              & Weight Memory: 2,048 KB   \\
\hline
  Maximum Input Image Size & VGA: $640 \times 480$ pixels\\
\hline
  Processing Speed & RetinaFace~\cite{Deng19}$^{3}$: 150 fps\\
                   & RetinaFace with DDC$^{3}$: 180 fps\\
\hline
  Size of Filter Kernels$^{4}$ & Maximum: $N_{\mathsf{ch},i} \times 7 \times 7$\\
\hline
  Performance &   512 MACs / clock cycle\\
\tabucline[1pt]{-}
\end{tabu}
\end{center}
{\small
$^{1}$CLPU stands for ``Convolution Layer Processing Unit."\\
$^{2}$APLPU stands for ``Activation and Pooling Layer Processing Unit."\\
$^{3}$The bit widths of filter weights and the feature maps are quantized to 8 bits. The post-processing time with an external CPU is included.\\
$^{4}N_{\mathsf{ch},i}$ represents the number of output channels in the $i$-th convolution layer in the network.\\
}
\caption{Specifications of the proposed embedded inference accelerator.}
\label{tab:spec}
\end{table}

Table~\ref{tab:utilization} shows the comparison of hardware specifications with three related works. Since it is difficult to compare the performance of FPGA and ASIC, we focus on the hardware utilization of MAC units for regular convolutions and depthwise convolutions. In the first related work~\cite{Liu19}, when dealing with depthwise convolutions, the filter kernels are filled with zeros, and the advantage is that both regular convolutions and depthwise convolutions can be computed by using the same hardware architecture. Suppose that the hardware can handle regular convolutions with $T_n$ input channels and $T_m$ output channels in parallel, the hardware utilization becomes $\frac{100}{T_m}\%$ when dealing with depthwise convolutions with $T_n$ input channels and $T_m$ output channels. In the second related work~\cite{Su18}, regular convolutions and depthwise convolutions are handled with different hardware architectures. The advantage is that both regular convolutions and depthwise convolutions can be handled efficiently, but some of the processing elements do not function when either depthwise convolutions or regular convolutions are handled. In the third related work~\cite{Yu20}, the processing elements can effectively handle regular convolutions and depthwise convolutions, but as discussed in Sec. \ref{subsec:time}, the performance decreases when the kernel size of filters is larger than or equal to $5 \times 5$.

The proposed architecture can handle regular convolutions and depthwise convolutions efficiently with large-kernel filters. The overhead to support the function is the buffer for input feature maps, which is mentioned in Sec.~\ref{subsec:network}. 

\begin{table}
\begin{center}
\begin{tabu}{lcccc}
\tabucline[1pt]{-}
          & Liu et al.~\cite{Liu19} &  Su et al.~\cite{Su18} & Yu et al.~\cite{Yu20} & This Work\\
\hline
  Device & ZYNQ7100 & XCZU9EG  & XCK325T & ASIC \\
  Frequency & 100MHz & 150MHz & 200MHz & 400MHz\\  
  Performance (GOPS) & 17.11 &  91.2 & 460.8 & 409.6\\
\hline
  Utilization:$^{1}$\\
  Regular Convolutions & 100\% & 100\% & 100\% & 100\%\\
  Depthwise Convolutions & 13\%$^{2}$ & 50\%$^{3}$ & $3 \times 3$ filters: 100\%  & 100\%\\
                         &            &            & $5 \times 5$ filters: 69\%$^{4}$ \\
\tabucline[1pt]{-}
\end{tabu}
\end{center}
{\small
$^{1}$Utilization represents the hardware utilization of MAC processing elements for regular convolutions and depthwise convolutions in an ideal case.\\
$^{2}$We suppose that $T_m = T_n = 8$~\cite{Liu19}, and the utilization becomes $100\% \times \frac{1}{T_m} = 13\%$. \\
$^{3}$We suppose that $M' = N'$~\cite{Su18}.\\
$^{4}$Since $\alpha = 4$~\cite{Yu20}, the utilization for $5 \times 5$ filter kernels is $100\% \times \left( \frac{5^2}{3^2 \cdot \alpha}  \right) = 69\%$.\\
}
\caption{Comparison with the related works.}
\label{tab:utilization}
\end{table}

\section{Conclusions and Future Work}
\label{sec:conclusion}

In this paper, a new solution which combines the advantages of algorithms and hardware architectures, is proposed to handle modern CNNs for embedded computer vision. The contribution of this paper is twofold. First, a new hardware architecture is proposed to handle both depthwise convolutions and regular convolutions. The proposed architecture can support filters with kernels larger than $3 \times 3$ with high efficiency, and the processing time is shorter than the related works. The experimental results show that the gate count of the proposed hardware is 1.97M gates, and the number of parallel processing units for MAC operations is 512. Second, the features of the supported network architecture, DDC, are analyzed with two applications. By replacing regular convolutions with large-kernel depthwise convolutions, it is possible to reduce the computational costs while keeping the accuracy.

For future work, we plan to extend the functions of the proposed hardware architecture to handle other kinds of complicated network architectures. In addition, we will apply the combination of the dilated convolutions and the depthwise convolutions to other applications.

\clearpage
%
%
\bibliographystyle{splncs04}
\bibliography{egbib}
\end{document}